\documentclass[letterpaper, 10 pt, conference]{ieeeconf}  
\IEEEoverridecommandlockouts                              
\usepackage{cite}

\usepackage{amsmath,amssymb,amsfonts}
\usepackage{algorithmic}
\usepackage{graphicx}
\usepackage{textcomp}
\usepackage{xcolor}
\usepackage{color}													
\usepackage{xfrac}														
\usepackage{colortbl}
\usepackage{siunitx}
\usepackage[english]{babel}	
\usepackage{multirow}
\usepackage{hyperref}
\usepackage{blindtext}
\usepackage{mathtools}
\usepackage{times}
\usepackage{epsfig}
\usepackage{bbding}
\usepackage{comment}

\definecolor{olivengruen}{RGB}{110, 117, 14}

\usepackage{subcaption}
\DeclareCaptionLabelFormat{andtable}{TABLE \thetable}

\title{\LARGE \bf
Skin the sheep not only once: \\ 
Reusing Various Depth Datasets to Drive the Learning of Optical Flow
}

\author{Sheng-Chi Huang \qquad Wei-Chen Chiu
\\
{\small\tt National Yang Ming Chiao Tung University, Hsinchu, Taiwan}
}

\begin{document}
	
    \graphicspath{{./graphics/}}
    \maketitle
    

    \begin{abstract}
        Optical flow estimation is crucial for various applications in vision and robotics. As the difficulty of collecting ground truth optical flow in real-world scenarios, most of the existing methods of learning optical flow still adopt synthetic dataset for supervised training or utilize photometric consistency across temporally adjacent video frames to drive the unsupervised learning, where the former typically has issues of generalizability while the latter usually performs worse than the supervised ones. To tackle such challenges, we propose to leverage the geometric connection between optical flow estimation and stereo matching (based on the similarity upon finding pixel correspondences across images) to unify various real-world depth estimation datasets for generating supervised training data upon optical flow. Specifically, we turn the monocular depth datasets into stereo ones via synthesizing virtual disparity, thus leading to the flows along the horizontal direction; moreover, we introduce virtual camera motion into stereo data to produce additional flows along the vertical direction. Furthermore, we propose applying geometric augmentations on one image of an optical flow pair, encouraging the optical flow estimator to learn from more challenging cases. Lastly, as the optical flow maps under different geometric augmentations actually exhibit distinct characteristics, an auxiliary classifier which trains to identify the type of augmentation from the appearance of the flow map is utilized to further enhance the learning of the optical flow estimator. Our proposed method is general and is not tied to any particular flow estimator, where extensive experiments based on various datasets and optical flow estimation models verify its efficacy and superiority.

    \end{abstract}

    \section{Introduction}

\begin{figure}[th!]
    \centering
    \includegraphics[width=\linewidth]{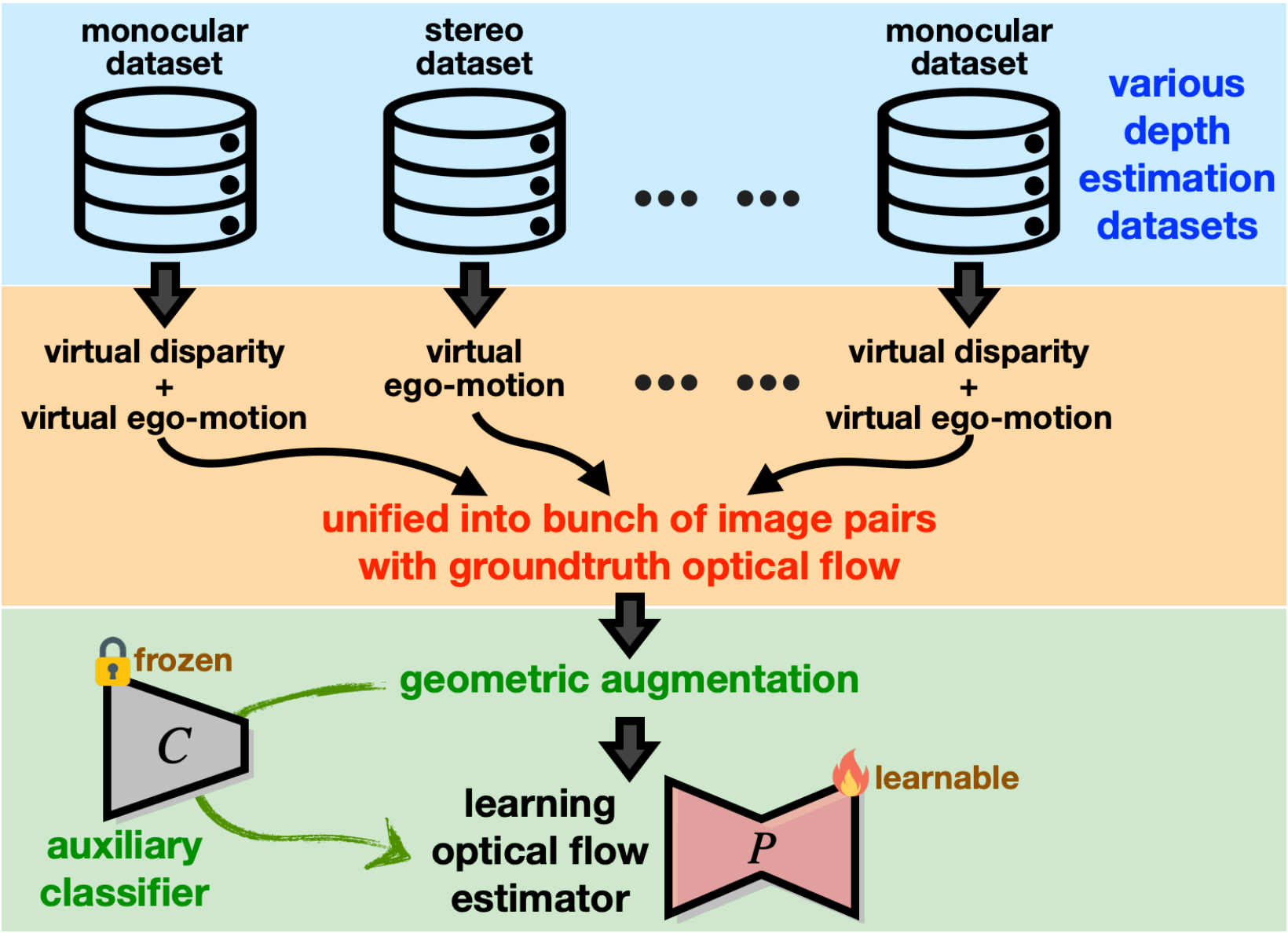}
    \caption{Our framework unifies various depth estimation datasets into supervised data of learning optical flow, via the introduction of virtual disparity and virtual camera motion (i.e. ego-motion) to produce the horizontal and vertical flows. In addition, the geometric augmentations are applied to not only generate more challenging data samples but also enable an auxiliary classifier for further boosting the training of optical flow estimator.}
    \label{fig:single-image geometric augmentation}
    \vskip -1.5em
\end{figure}

Optical flow estimation plays an important role across plenty applications such as robotics, augmented reality, and autonomous vehicles. Although there exist many traditional approaches~\cite{lucas1981iterative,brox2004high,baker2011database,menze2015discrete,chen2016full} which attempt to model such a problem of finding dense pixel-wise displacement across images from different perspectives, their optimization objectives are typically hand-crafted thus being hard to handle various corner cases. Along with the recent advance of deep learning techniques, we have witnessed the magic leap on performance for optical flow estimation brought by deep neural networks (e.g. FlowNet~\cite{dosovitskiy2015flownet} as the seminal work and many others \cite{ranjan2017optical,sun2018pwc,teed2020raft,zhang2021separable,xu2022gmflow}). While most of these works heavily rely on large-scale datasets with groundtruth annotations (i.e. optical flow maps) to perform the supervised learning, collecting such supervised datasets is highly challenging and expensive in the real world (as there exists no sensor which can directly measure the pixel-wise correspondence between views). To address this problem, many research works \cite{ilg2017flownet, ranjan2017optical, hui2018liteflownet, sun2018pwc} have utilized a pre-training approach on large synthetic datasets \cite{ilg2017flownet, mayer2016large}, followed by fine-tuning on limited target datasets \cite{butler2012naturalistic, geiger2012we, menze2015object}. However, such an approach is still hampered by a lack of groundtruth in the real world (for performing fine-tuning) and suffers from poor generalizability due to the domain shift (i.e. different distributions between synthetic and real-world data). Although there exist several recent attempts to explore the unsupervised learning scenario \cite{meister2018unflow,zou2018df,liu2019selflow, lai2019bridging} (where the photometric consistency is usually adopted to evaluate pixel correspondences across images, and joint learning with other tasks such as depth estimation or camera pose estimation would come into play), their performances are mostly still inferior to those supervised ones. 



To strike a better balance among all the aforementioned challenges (e.g., lack of real-world supervised dataset, domain shift, and inferior performance for unsupervised learning), \cite{aleotti2021learning} recently proposed a novel method to generate an image pair with its optical flow annotations from a single real-world input image. Basically, given the input image, its corresponding 3D point cloud is firstly built with the help of depth estimation (i.e. projecting each pixel in the input image back to 3D space), then a 3D motion (composed of rotation and translation) is applied to the virtual camera of the input image to synthesize a novel view, where the pixel correspondence (i.e. optical flow map) between the original input image and the novel view is naturally available as the entire geometric transformation and projection procedure is under manual control. In the results, the real-world supervised dataset can be constructed for training the optical flow estimator. 

Inspired by~\cite{aleotti2021learning}, we come up with two further considerations: 1) While the accuracy of depth estimation is critical to the quality of synthesized novel views in ~\cite{aleotti2021learning}, the monocular depth estimation datasets which contain groundtruth depth maps for their images seem to be a feasible alternative to bypass the uncertainty stemmed from depth estimation; 2) As optical flow estimation has a close relative, i.e. stereo matching, in terms of finding correspondences, the stereo depth estimation datasets ideally ought to be helpful as well for learning optical flow. Moreover, as collecting groundtruth depth maps in real-world scenarios is typically more achievable than the optical flow ones due to the popularity of depth cameras, we are therefore motivated to bring up the following question: \emph{Can optical flow estimation be learned from both monocular and stereo depth datasets, and can the relationship between stereo matching and optical flow be explored beyond treating depth as an intermediate product?}

To this end, we propose a framework to unify both monocular and stereo depth datasets, followed by transforming them into a collection of annotated optical flow data: For an image in the monocular depth dataset, we first translate its groundtruth depth map into the disparity map (named \textbf{virtual disparity}), which is used to warp the original image into a novel view, where the original image and the warped one together become a stereo image pair. It is worth noting that the pixel correspondence in such stereo image pair only has the horizontal offset, which can be treated as horizontal optical flow; while for a stereo image pair (obtained from the stereo depth dataset or produced by the previous step), we can apply the same procedure as~\cite{aleotti2021learning} to employ virtual camera motion (also named as \textbf{virtual ego-motion}) on one image of such pair, in which its resultant novel view together with the other image of the stereo pair finally form an optical flow pair containing groundtruth annotation.

Furthermore, as data augmentation has become a widely adopted training strategy to increase the quantity and diversity of training data for further improving the performance of deep models, most works of optical flow estimation also employ data augmentation, in which the used augmentation operations can be categorized into two classes: 1) \textbf{photometric augmentations}, which mainly modify the pixel appearance (e.g. contrast, sharpness, brightness, and colors) while preserving the spatial structure, where such property allows them to be applied independently to each image; 2) \textbf{geometric augmentations}, which would affect the scene structure (e.g. flipping, cropping, rotating, and scaling), thus they should be applied to both images in a pair and the corresponding optical flow map simultaneously.

In this work, we step further to discover that, once we only perform the geometric augmentations on one image in a pair, its resultant pixel correspondence with respect to another image (i.e. optical flow map) will undergo a drastic distortion and becomes more challenging for the optical flow estimator to predict, thus including such data samples into training would benefit the learning of optical flow estimation model. Moreover, we have another observation that applying different geometric augmentations on only one image in a pair would lead to optical flow maps with distinct characteristics. In the results, we propose to pretrain an auxiliary classifier which takes the optical flow map as input and predicts the type of geometric augmentation being applied, where such a classifier can be later utilized to construct a novel objective for providing additional supervision signals in training optical flow estimator. In summary, our full method is composed of all the designs above (i.e. unification over various depth estimation datasets to create a supervised optical flow dataset, including challenging training samples via applying geometric augmentations on only one image in an optical flow pair, and using the pretrained auxiliary classifier to boost the optical flow learning), in which it can be employed on training any optical flow estimator thus being quite flexible and general. 
    \section{Related Work}

\begin{figure*}[ht]
    \begin{minipage}[t]{0.64\linewidth}
        \centering
        \includegraphics[width=\linewidth]{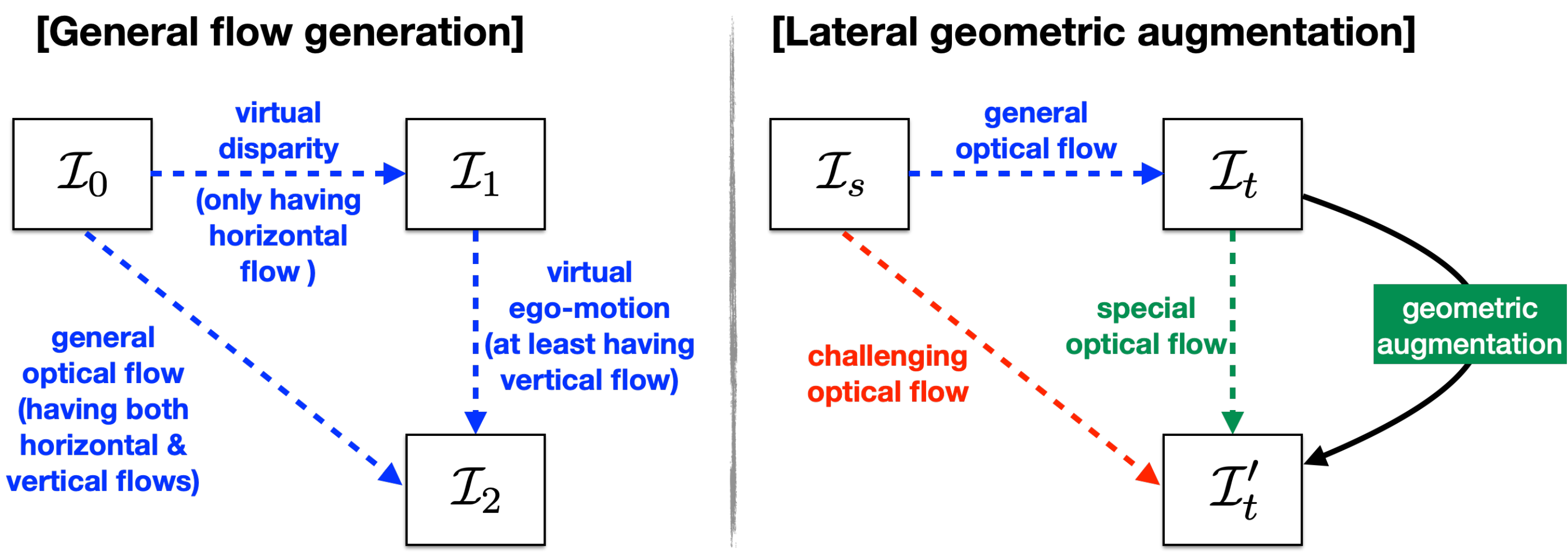}
        \vspace{-1.75em}
        \caption{\textbf{Overview of general flow generation (left figure, cf. Section~\ref{sec:general_flow}) and lateral geometric augmentation (right figure, cf. Section~\ref{sec:lateral}.} In general flow generation, virtual disparity is firstly adopted to turn an image from the monocular depth dataset into a stereo pair, then the virtual ego-motion is applied on one image of each stereo pair (regardless from monocular or stereo datasets) to include vertical flows, finally we can produce the optical flow training sample with general flow map (i.e. having both horizontal and vertical flows); Our lateral geometric augmentation applies geometric augmentation on only one image of a training pair to produce more challenging optical flow cases.}
        \label{fig:preprocess and augmentation}
    \end{minipage}%
        \hfill%
    \begin{minipage}[t]{0.34\linewidth}
        \centering
        \includegraphics[width=\linewidth]{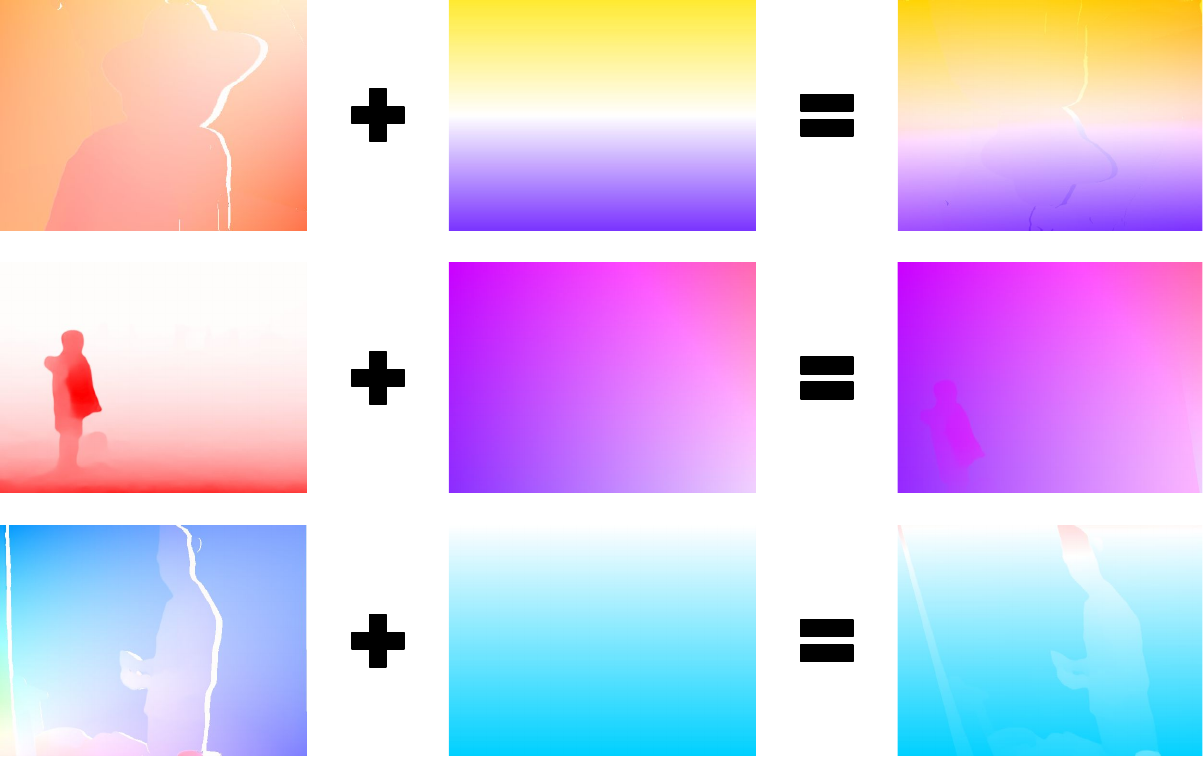}
        \caption{\textbf{Examples of distinct optical flow characteristics from various lateral geometric augmentations (cf. Section~\ref{sec:lateral}).} From top to bottom (zoom in for better view), the rightmost flows are generated by applying flipping, rotation, and shearing operations on the leftmost source flows (following Eq.~(\ref{eq:cflow-1}) and ~(\ref{eq:cflow-2})).}
        \label{fig:features}
    \end{minipage}

    \vskip -1.5em
    
\end{figure*}

In recent years, there has been a growing interest in leveraging the geometric relationship between stereo matching (or depth estimation) and optical flow estimation to jointly learn spatial correspondence and improve flow accuracy \cite{zou2018df, lai2019bridging, chi2021feature, guizilini2022learning}. For instance,~\cite{lai2019bridging} finds that the optical flow estimation and stereo matching tasks co-exist in the temporally adjacent stereo pairs (i.e., stereo video) thus proposes to bridge both tasks via photometric reconstruction in such data format for learning a unified model of finding pixel correspondence across images. Instead of jointly learning optical flow and depth estimation, some works propose to use depth estimation as an intermediate step to generate the training data for learning estimation of pixel correspondence (e.g. stereo matching or optical flow estimation). For instance, \cite{watson2020learning} and \cite{aleotti2021learning} adopt the estimated depth of the input image to perform novel view synthesis for, respectively, constructing the annotated stereo or optical flow pairs. Our proposed framework follows the similar idea as~\cite{aleotti2021learning} but directly takes the advantage of using groundtruth depth in the supervised depth estimation datasets. Moreover, as our framework has the feature of unifying various depth datasets, it is also conceptually related to \cite{ranftl2022midas}, in which \cite{ranftl2022midas} allows the mixture of multiple datasets with various formats of depth annotations during training the depth estimator. Nevertheless, our target scenario is quite different from \cite{ranftl2022midas}.
    \section{Methodology}

\subsection{Create Annotated Flow via Unifying Depth Datasets}
    \label{sec:general_flow}
    \noindent \textbf{Producing horizontal optical flow.}
    As previously motivated, we would like to unify various real-world depth estimation datasets with groundtruth annotations, regardless of the monocular or stereo ones, for synthesizing the supervised dataset for learning optical flow estimation. We denote an optical flow training sample as a tuple $(\mathcal{I}_0, \mathcal{I}_1, F_{0 \rightarrow 1})$, where $\mathcal{I}_0$ and $\mathcal{I}_1$ form an image pair, while $F_{0 \rightarrow 1}$ represents the optical flow map between $\mathcal{I}_0$ and $\mathcal{I}_1$.

    Given an image $\mathcal{I}_0^{m}$ and its corresponding groundtruth depth map $Z_0^{m}$ obtained from a monocular depth dataset, we start by converting $Z_0^{m}$ into the disparity map $\Tilde{d}_{m}$ via 
        $\Tilde{d}_{m} = \frac{B f}{Z_0^{m}}$
    where $B$ and $f$ denote the baseline and focal length, respectively. Following the practice in~\cite{watson2020learning}, in order to have a wide range of baselines and focal lengths, we set $Bf=s_c$ in which such a scaling factor $s_c$ is randomly drawn from a uniform distribution. As the disparity represents the pixel-wise displacement along the horizontal direction, we can easily translate it into the form of an optical flow map by $\left \langle \Tilde{d}_{m}, \varnothing \right \rangle$, where $\left \langle \cdot \right \rangle$ denotes the channel-wise concatenation operation, and $\varnothing$ is a zero map of the same size as $\Tilde{d}_{m}$. In particular, we can treat $\left \langle \Tilde{d}_{m}, \varnothing \right \rangle$ as a horizontal optical flow $F_{0 \rightarrow 1}^{m}$. We can now construct a stereo pair $\{\mathcal{I}_0^{m}, \Tilde{\mathcal{I}}_1^{m}\}$ via  
    \begin{align}
        \Tilde{\mathcal{I}}_1^{m} &= \mathcal{W}(\mathcal{I}_0^{m}, s_i \cdot F_{0 \rightarrow 1}^{m}) \\
        \Tilde{Z}_1^{m} &= \mathcal{W}(Z_0^{m}, s_i \cdot F_{0 \rightarrow 1}^{m})
    \end{align}
    where $\mathcal{W}(\alpha, \beta)$ defines the warping function in which $\alpha$ is warped according to $\beta$, $s_i$ is randomly set to either $1$ or $-1$ to simulate the respective case where $\mathcal{I}_0^{m}$ is on the left- or right-hand side, and $\Tilde{Z}_1^{m}$ is the depth map for $\Tilde{\mathcal{I}}_1^{m}$.

    Next, we consider an image pair $\{\mathcal{I}_0^{s}, \mathcal{I}_1^{s}\}$ obtained from a stereo dataset, where the corresponding groundtruth disparity map is $d_{s}$. The depth map $\Tilde{Z}_0^{s}$ of $\mathcal{I}_0^{s}$ can be simply computed by 
        $\Tilde{Z}_0^{s} = \frac{B f}{d_{s}}$
    where $Bf$ is set to a constant as $d_s$ contains actual disparity values. And the depth map $\Tilde{Z}_1^{s}$ of $\mathcal{I}_1^{s}$ is:
    \begin{equation}
        \Tilde{Z}_1^{s} = \mathcal{W}(\Tilde{Z}_0^{s}, F_{0 \rightarrow 1}^{s}) \text{,~where~} F_{0 \rightarrow 1}^{s} = \left \langle d_{s}, \varnothing \right \rangle.
    \end{equation}

    \noindent \textbf{Synthesizing general optical flow.}
    Since $F_{0 \rightarrow 1}^{m}$ or $F_{0 \rightarrow 1}^{s}$ only have the horizontal displacement, now we step further to utilize virtual camera motion (i.e. virtual ego-motion) as~\cite{aleotti2021learning} to produce more general optical flow maps (i.e. having both horizontal and vertical displacements). Taking the stereo pair $\{\mathcal{I}^{s}_0, \mathcal{I}^{s}_1\}$ and the corresponding flow map $F^{s}_{0 \rightarrow 1}$ as an example, we aim to create a plausible optical flow $F^s_{1 \rightarrow 2}$ that at least has vertical displacement. This operation will enable synthesizing a novel view $\Tilde{\mathcal{I}}^s_2$, its corresponding depth map $\Tilde{Z}^s_2$, and the target general optical flow $F^s_{0 \rightarrow 2}$ (i.e. the flow map from $\mathcal{I}^{s}_0$ to $\Tilde{\mathcal{I}}^s_2$).

    Basically, we first hypothesize an intrinsic matrix $K$ in which its inverse $K^{-1}$ will be used to project the pixels in $\mathcal{I}^s_1$ back into the 3D space, following the common practice of previous approaches \cite{yin2018geonet, guizilini2022learning, lai2019bridging}. Then we randomly sample a plausible rotation $R_{1 \rightarrow 2}$ and translation $t_{1 \rightarrow 2}$ to obtain a transformation matrix $T_{1 \rightarrow 2} = \left [ R_{1 \rightarrow 2} \mid t_{1 \rightarrow 2} \right]$. Subsequently, we project each pixel $p_1 \in \mathcal{I}^s_1$ back into 3D space to form a point cloud, perform camera ego-motion, and finally project the 3D point cloud onto the 2D image plane to derive the optical flow map $F^s_{1 \rightarrow 2}$. With such an optical flow map $F^s_{1 \rightarrow 2}$, we can then synthesize the novel view $\Tilde{\mathcal{I}}^{s}_2$ and its corresponding depth map $\Tilde{Z}^{s}_2$.

    \begin{figure*}[ht]
        \centering
        \includegraphics[width=.92\textwidth]{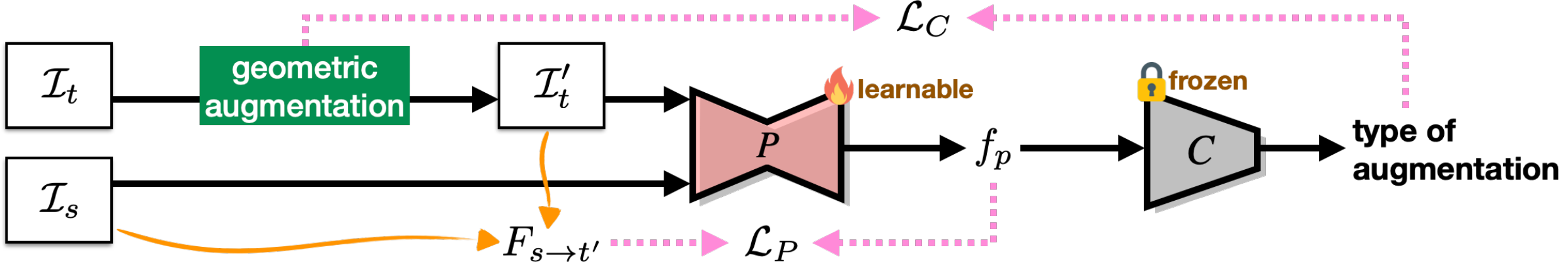}
        \caption{\textbf{Our framework for learning optical flow estimator.} Given an image pair $\{\mathcal{I}_s, \mathcal{I}_t\}$, we firstly apply lateral geometric augmentation on $\mathcal{I}_t$ to obtain $\mathcal{I}'_t$, where the resultant groundtruth optical flow $F_{s \rightarrow t'}$ between $\mathcal{I}_s$ and $\mathcal{I}'_t$ can be easily obtained via the computation in Section~\ref{sec:lateral}. Our flow estimator $P$ which takes $\{\mathcal{I}_s, \mathcal{I}'_t\}$ as input is trained to output the flow map $f_p$ that ideally should be identical to $F_{s \rightarrow t'}$, where the objective $\mathcal{L}_P$ evaluates the difference between $F_{s \rightarrow t'}$ and $f_p$. Moreover, the auxiliary classifier $C$, which is trained to identify the type of augmentation used in the lateral geometric augmentation of the appearance of input flow, contributes to define another objective $\mathcal{L}_C$ (cf. Section~\ref{sec:classifier}) to increase the learning of $P$.}
        \label{fig:training framework}
        \vskip -1.5em
    \end{figure*}

    The overall procedure to reach $F^{s}_{0 \rightarrow 2}$ is summarized as:
    \begin{align}
        F^s_{1 \rightarrow 2} &= K T_{1 \rightarrow 2} \Tilde{Z}^s_1(p_1) K^{-1} p_1 - p_1 \\
        \Tilde{\mathcal{I}}^{s}_2 &= \mathcal{W}(\mathcal{I}^s_1, F^s_{1 \rightarrow 2}), 
        \Tilde{Z}^{s}_2 = \mathcal{W}(\Tilde{Z}^s_1, F^s_{1 \rightarrow 2}) \\
        F^{s}_{0 \rightarrow 2} &= F^s_{0 \rightarrow 1} + \mathcal{W}^{-1}(F^s_{0 \rightarrow 1}, F^s_{1 \rightarrow 2})
    \end{align}
    where $\mathcal{W}^{-1}(\alpha, \beta)$ is a specific warping function to support the backward warping that $\mathcal{W}^{-1}(\alpha, \beta) = \beta(x + \alpha(x))$ and $x$ denote all pixel locations. Note that we can apply the same procedure to the stereo pair $\{\mathcal{I}_0^{m}, \Tilde{\mathcal{I}}_1^{m}\}$ derived from the monocular dataset to obtain $F^{m}_{0 \rightarrow 2}$.

    Finally, we collect all the produced image pairs and their corresponding groundtruth optical flow maps, in which they are represented as tuples: $(\mathcal{I}^m_0, \Tilde{\mathcal{I}}^{m}_1, F^m_{0 \rightarrow 1})$, $(\tilde{\mathcal{I}}^{m}_1, \Tilde{\mathcal{I}}^{m}_2, F^m_{1 \rightarrow 2})$, $(\mathcal{I}^m_0, \tilde{\mathcal{I}}^{m}_2, F^m_{0 \rightarrow 2})$, $(\mathcal{I}^s_0, \mathcal{I}^{s}_1, F^s_{0 \rightarrow 1})$, $(\mathcal{I}^{s}_1, \Tilde{\mathcal{I}}^{s}_2, F^s_{1 \rightarrow 2})$ and $(\mathcal{I}^s_0, \Tilde{\mathcal{I}}^{s}_2, F^{s}_{0 \rightarrow 2})$. The left portion of Figure \ref{fig:preprocess and augmentation} shows the entire aforementioned procedure to create optical flow training samples, which is named \textbf{general flow generation}.

\subsection{Lateral Geometric Augmentation}
    \label{sec:lateral}
    \noindent In addition to typical augmentation strategies (i.e. applying photometric augmentation on each image independently, or applying the same geometric augmentation on the two images of a training pair), we propose to conduct \textbf{lateral geometric augmentation}, where the geometric augmentation is applied on only one image of each training pair to produce more challenging cases to learn the optical flow estimation. Three three geometric augmentation operations are adopted and sequentially introduced in the following. 
    \noindent \textbf{Flipping operation.}
    Both horizontal and vertical flipping operations are adopted.
    After applying the horizontal flipping operation, we use the equations below, which are derived from the symmetric attribution of the flipped coordinates.
    \begin{equation}
        (x_0 + x_1)/2 = H/2, \quad y_0 = y_1
    \end{equation}
    where $p_0 = (x_0, y_0)$ and $p_1 = (x_1, y_1)$ represent the original and flipped pixel coordinates, respectively, and $H$ is the height of the image, we reach the per-pixel flow $f^{hf}_{p_0 \rightarrow p_1} = (H - 2x_0, 0)$ for horizontal flipping augmentation. Similarly, for the vertical flipping operation, we obtain the per-pixel flow $f_{p_0\rightarrow p_1}^{vf} = (0, W - 2y_0)$ where $W$ is the width of the image. When collecting all the per-pixel flow, we can obtain the special flow $F_a^{hf}$ and $F_a^{vf}$ related to horizontal and vertical flipping augmentations, respectively. In particular, the backward flow $B$ resulting from the flipping augmentation is the same as the forward flow $F$, so we have $B_a^{hf} = F_a^{hf}$ and $B_a^{vf} = F_a^{vf}$.

    \noindent \textbf{Rotation operation.}
    We apply rotation operations to images by randomly sampling a Euler angle $\theta_a$ and adding a sign factor $s_i$ uniformly sampled from $\{-1, 1\}$ to simulate clockwise and counterclockwise rotations. The rotation matrix $R({\theta_a})$ is then calculated by:
    \begin{equation}
        R(s_i{\theta_a}) = \left [
            \begin{matrix}
                cos(s_i \theta_a) & -sin(s_i \theta_a) \\
                sin(s_i \theta_a) & cos(s_i \theta_a) \\
            \end{matrix}
        \right ]
    \end{equation}
    Next, we randomly sample a center point $c_0$ and rotate every pixel coordinate $p_0$ about $c_0$ by applying the rotation matrix. This yields the special flow $F_a^{r}$ and the backward flow $B_a^{r}$ resulting from the rotation augmentation: 
    \begin{align}
    F_a^r &= R(s_i\theta_a) (p_0 - c_0) + c_0 - p_0 \\
    B_a^r &= R(-s_i\theta_a) (p_0 - c_0) + c_0 - p_0
    \end{align}
    \noindent \textbf{Shearing operation.}
    We introduce shearing operations to images along horizontal or vertical directions by randomly sampling a shearing magnitude factor $\lambda_a$ and a sign factor $s_i$ that is sampled uniformly from $\{-1, 1\}$ to simulate different directions of shearing stress. This creates horizontal and vertical shearing matrices $S^{hs}(\lambda_a)$ and $S^{vs}(\lambda_a)$, respectively:
    \begin{equation}
        S^{hs}(\lambda_a) = \left [
            \begin{matrix}
                1 & s_i \lambda_a \\
                0 & 1 \\
            \end{matrix}
        \right ],
        S^{vs}(\lambda_a) = \left [
            \begin{matrix}
                1 & 0 \\
                s_i \lambda_a & 1 \\
            \end{matrix}
        \right ]
    \end{equation}
    We then apply these horizontal or vertical shearing matrices to every pixel coordinate $p_0$ to obtain new coordinates $p_1$. To obtain the forward flows $\{F_a^{hs}, F_a^{vs}\}$ and backward flows $\{B_a^{hs}, B_a^{vs}\}$ stemmed from shearing augmentation, we use:
    \begin{align}
            F_a^{hs} &= S^{hs}(\lambda_a) p_0 - p_0, \quad
            B_a^{hs} = S^{hs}(-\lambda_a) p_0 - p_0 \\
            F_a^{vs} &= S^{vs}(\lambda_a) p_0 - p_0, \quad
            B_a^{vs} = S^{vs}(-\lambda_a) p_0 - p_0
    \end{align}

    \noindent \textbf{Lateral geometric augmentation.}
    Given a tuple of training data $(\mathcal{I}_{s}, \mathcal{I}_{t}, F_{s \rightarrow t})$ which can be mapped to any tuple produced by general flow generation (cf. Section~\ref{sec:general_flow}), we can apply geometric augmentation to either the source $\mathcal{I}_{s}$ or target image $\mathcal{I}_{t}$ to obtain $\mathcal{I}'_{s} = A(\mathcal{I}_s)$ or $\mathcal{I}'_{t} = A(\mathcal{I}_t)$, where $A(\cdot)$ is the augmentation operator. We refer to the forward flow resulting from flipping, rotation, or shearing augmentation as $F_a$ (also named as \textit{special flow}), which is used to warp $\mathcal{I}_s$ to $\mathcal{I}'_{s}$, and the backward flow as $B_a$, which is used to warp $\mathcal{I}'_{t}$ to $\mathcal{I}_t$. We then use the warping operator $\mathcal{W}^{-1}$ to compute the challenging flow cases:
    \begin{align}
        \label{eq:cflow-1}
        F_{s \rightarrow t'} &= F_{s \rightarrow t} + \mathcal{W}^{-1}(F_{s \rightarrow t}, F_a) \\
        \label{eq:cflow-2}
        F_{s' \rightarrow t} &= B_{a} + \mathcal{W}^{-1}(B_{a}, F_{s \rightarrow t})
    \end{align}
    The resulting tuples of $(\mathcal{I}_{s}, \mathcal{I}'_{t}, F_{s \rightarrow t'})$ and $(\mathcal{I}'_{s}, \mathcal{I}_{t}, F_{s' \rightarrow t})$, generated by lateral geometric augmentation, are more challenging and allow us to train the optical flow estimation network more effectively. The right portion of Figure \ref{fig:preprocess and augmentation} shows an overview of our \textbf{lateral geometric augmentation}.
    
    \begin{figure*}[ht]
        \centering
        \includegraphics[width=\textwidth]{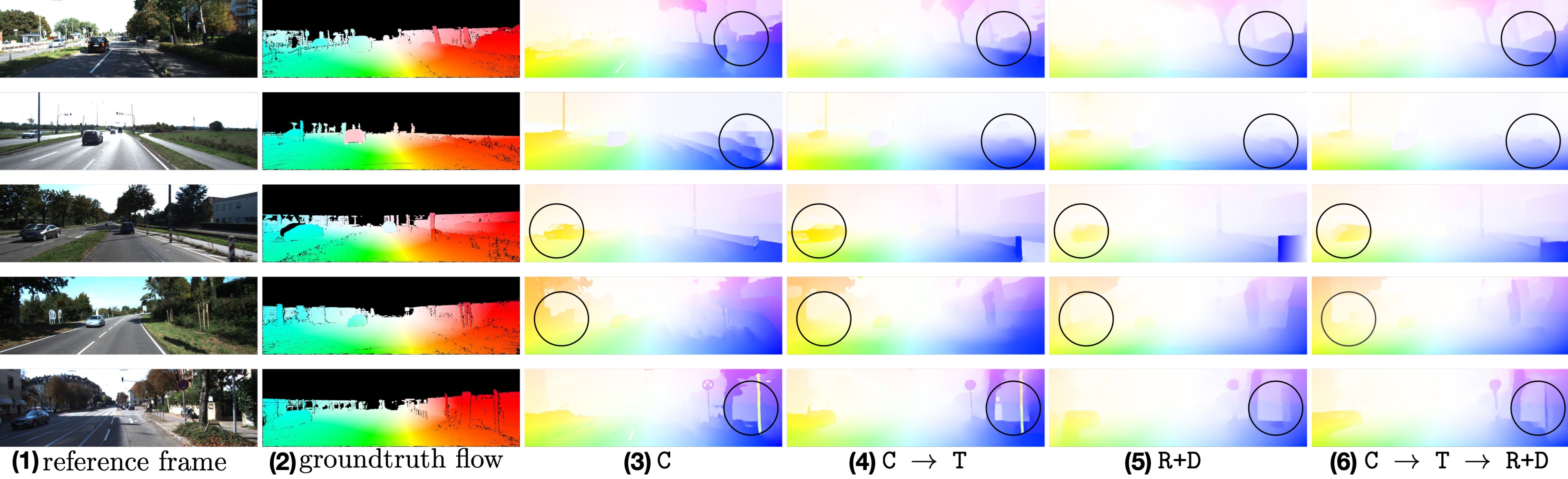}
        \vspace{-1.5em}
        \caption{\textbf{Example Results on KITTI-15}: (1) reference frame, (2) groundtruth flow, and flow maps produced by the RAFT models respectively trained on (3) FlyingChairs (\texttt{C}), (4) FlyingChairs$\rightarrow$FlyingThings3D (\texttt{C$\rightarrow$T}), (5) our real-world dataset mixed across ReDWeb and DIML (\texttt{R+D}), and (6) FlyingChairs$\rightarrow$FlyingThings3D$\rightarrow$our mixed dataset (\texttt{C$\rightarrow$T$\rightarrow$R+D}). }
        \label{fig:kitti results}
        \vskip -1.5em
    \end{figure*}

\subsection{Auxiliary Classifier}\label{sec:classifier}
    \noindent \textbf{Optical flow features from different augmentations.} 
    As shown by Eq.~(\ref{eq:cflow-1}) and~(\ref{eq:cflow-2}), both optical flow maps $F_{s \rightarrow t'}$ and $F_{s' \rightarrow t}$ contain the augmentation-dependent components of forward flow $F_a$ and backward flow $B_a$, it is thus believed that we should be able to infer the type of augmentation (flipping, rotation, or shearing) being applied for the lateral geometric augmentation. We achieve this by training an auxiliary classifier $C$ that can identify the type of augmentation used in lateral geometric augmentation from the appearance of the input optical flow map. The classifier takes an optical flow $F$ as input and produces a real-valued vector $T_p = C(F)$ composed of four elements, where $T_p$ passed through softmax represents the posterior of four augmentation types (i.e. flipping, rotation, shearing, and none of the above). Our classifier $C$ comprises a feature extractor, an average pooling layer, and a fully connected layer. The feature extractor is based on the small encoder of the RAFT \cite{teed2020raft} model with input channels set to 2 to support the optical flow input.

    \noindent\textbf{Training optical flow estimator with auxiliary classifier.}
    With the auxiliary classifier $C$ pretrained on annotated optical flow data (produced by general flow generation and later geometric augmentation), we can further use it to increase the learning of the flow estimator $P$, as shown in Figure \ref{fig:training framework}. Given an optical flow training pair $\{\mathcal{I}_s, \mathcal{I}_t\}$, we apply lateral geometric augmentation on $\mathcal{I}_t$ with the augmentation type $T_a$ to produce $\mathcal{I}'_t$. When the flow estimator $P$ takes $\{\mathcal{I}_s, \mathcal{I}'_t\}$ as input, its prediction $f_p$ should not only align with the groundtruth flow map $F_{s \rightarrow t'}$ (where the L1 error between $f_p$ and $F_{s \rightarrow t'}$ defines the objective $\mathcal{L}_P$), but also to be correctly classified by $C$ to match $T_a$. The classification difference in terms of cross entropy between $C(f_p)$ and $T_a$ thus forms a novel objective $\mathcal{L}_{C}$ where the gradients are backpropagated to update $P$. The idea here is that since our classifier $C$ is pretrained and frozen during the training of flow estimator $P$, if the optical flow map $f_p$ predicted by $P$ is not accurate enough for the classifier $C$ to distinguish the type of augmentation being applied on $\mathcal{I}'_t$ then the error is all attributed to the model $P$. The overall loss $\mathcal{L}$ to train the flow estimator $P$ is formulated as:
    \begin{equation}
        \mathcal{L} = \mathcal{L}_P(f_p, F_{s \rightarrow t'}) + \lambda_{C} \mathcal{L}_C(C(f_p), T_a)
    \end{equation}
    where the hyperparameter $\lambda_C$ balances between $\mathcal{L}_P$ and $\mathcal{L}_C$.

    \section{Experiments}

\subsection{Datasets}\label{sec:train_dataset}
    Our approach enables the generation of annotated optical flow data via unifying a variety of depth estimation datasets, where two real-world depth datasets are used in our experiments: \textbf{ReDWeb} \cite{xian2018monocular} (a monocular dataset, denoted as \texttt{R}) and \textbf{DIML} (a stereo dataset, denoted as \texttt{D}). For detailedness, RebWeb \texttt{R} is an RGB-D dataset composed of 3600 images and their corresponding depth maps, moreover, it contains highly diverse indoor and outdoor scenes; DIML \texttt{D} focuses on outdoor scenes, in which it is composed of 1505 stereo pairs and the corresponding disparity maps.

    Moreover, as the synthetic datasets, which have groundtruth optical flow maps to enable supervised learning of optical flow estimation, are actually compatible with our proposed framework (cf. Figure~\ref{fig:training framework}), we therefore also consider them in our experiments: \textbf{FlyingChairs} (denoted as \texttt{C}) and \textbf{FlyingThings3D} (denoted as \texttt{T}), where \texttt{T} offers more complex motion patterns than those in \texttt{C}.

    Based on the four aforementioned datasets, we come up with several settings to leverage them for our training:
    \begin{itemize}
        \item \texttt{C} where the flow estimator $P$ is trained on \texttt{C} only.
        \item \texttt{C$\rightarrow$T} where the flow estimator $P$ is firstly trained on \texttt{C} then finetuned on the more complex dataset \texttt{T}.
        \item \texttt{R+D} where the flow estimator $P$ is trained on the real-world annotated optical flow dataset, which is produced by unifying/mixing across \texttt{R} and \texttt{D} datasets via our general flow generation procedure (cf. Section~\ref{sec:general_flow}).
        \item \texttt{C$\rightarrow$T$\rightarrow$R+D} where the flow estimator $P$ is sequentially trained/fine-tuned on dataset \texttt{C}, dataset \texttt{T}, and our mixed real-world dataset \texttt{R+D}. 
    \end{itemize}
    Furthermore, as our proposed framework for learning optical flow has the flexibility to support arbitrary backbones of flow estimator, here we adopt two well-known optical flow models for $P$ in our experiments: \textbf{RAFT}~\cite{teed2020raft} and \textbf{GMFlow}~\cite{xu2022gmflow}.   
    

    One synthetic and one real-world datasets are adopted for our evaluation: \textbf{Sintel} \cite{butler2012naturalistic} and \textbf{KITTI} \cite{geiger2012we, menze2015object}, where both are popular and challenging benchmarks for optical flow estimation. Sintel is generated from a 3D animated short film, which offers two render passes: ``clean'' and ``final'', where the latter additionally includes visual variations (e.g. blurring and atmospheric effects); KITTI is collected from real-world street views, where we use both versions of it, namely ``KITTI-12'' and ``KITTI-15''. Regarding the evaluation metrics, we adopt \textbf{EPE} and \textbf{F1-all}, where the former refers to the average endpoint error while the latter refers to the percentage of optical flow outliers over all pixels.

\subsection{Experimental Results}
    The experimental results of learning flow estimators (i.e. RAFT or GMFlow) upon four training settings are summarized in Table \ref{table:raft results} (for using RAFT model as flow estimator) and Table \ref{table:gmflow results} (for using GMFlow as flow estimator) respectively. From these results we draw several observations:\\ 
    \textbf{1)} Regarding the evaluation upon real-world scenarios (i.e. KITTI-12 and KITTI-15), training using real-world datasets \texttt{R+D} (where the optical flow annotations in \texttt{R+D} are produced by our proposed general flow generation), either only using \texttt{R+D} or starting from synthetics ones \texttt{C$\rightarrow$T} then finetuning on \texttt{R+D} (i.e. last two rows in both tables), provide better performance than those using only synthetic datasets (i.e. \texttt{C} and \texttt{C$\rightarrow$T}), showing our main contribution of providing real-world supervised dataset for optical flow estimation;\\
    \textbf{2)} Leveraging synthetic dataset that has precise optical flows of groundtruth to warm start the training (i.e. \texttt{C$\rightarrow$T$\rightarrow$R+D}) often is able to perform the best in real-world scenarios (i.e. KITTI), compared to using only the real-world dataset \texttt{R+D};\\
    \textbf{3)} Though having better performance and generalizability in real-world scenes (i.e. KITTI-12 and KITTI-15), training or finetuning on real-world datasets (i.e. \texttt{R+D} and \texttt{C$\rightarrow$T$\rightarrow$R+D}) would lead to worse performance in the synthetic scenario (i.e. Sintel) due to the domain gap among real and virtual data, while training ended up with synthetic ones (i.e. \texttt{C} and \texttt{C$\rightarrow$T}) typically performs better in Sintel, since \texttt{C} and \texttt{T} datasets share the same graphics rendering process as Sintel. Note that \cite{aleotti2021learning} also has such an observation similar to ours;\\
    \textbf{4)} Another important contribution of our proposed method is the capability of unifying various depth estimation datasets, where prior work \cite{aleotti2021learning} typically can only utilize a single dataset, thus resulting in worse performance (in which the model variant without any of our proposed designs, i.e. the first row in Table~\ref{table:DIML ablations}, actually reproduces \cite{aleotti2021learning}'s model trained on \texttt{D}; even when we train \cite{aleotti2021learning}'s model based on RAFT flow estimator upon \texttt{R}, it still performs worse to have 2.42 EPE and 9.96 F1-all for KITTI-12 and 5.65 EPE and 18.85 F1-all for KITTI-15) than ours trained on unified dataset \texttt{R+D}.

        \begin{table}[h!]
        \vspace{-.5em}
        \caption{ \textbf{Quantitative results upon RAFT flow estimators with different training settings (cf. Section~\ref{sec:train_dataset}).}}
        \label{table:raft results}
        \vspace{-.5em}
        \centering
        \begin{tabular}{ccccccccccc}
            \hline
            \multirow{2}{*}{Datasets} & \multicolumn{2}{c}{Sintel} & \multicolumn{2}{c}{KITTI-12} & \multicolumn{2}{c}{KITTI-15} \\
            \cline{2-3} \cline{4-5} \cline{6-7}
              & clean & final & EPE & F1-all & EPE & F1-all \\
            \hline
            \texttt{C} & \underline{2.36} & 4.39 & 5.14 & 34.64 & 10.77 & 41.08 \\
            \texttt{C $\rightarrow$ T} & \textbf{1.64} & \textbf{2.83} & 2.40 & 10.49 & 5.62 & 18.71 \\
            \texttt{R+D} & 2.61 & 3.98 & \underline{2.16} & \underline{9.28} & \underline{4.18} & \underline{15.03} \\
            \texttt{C $\rightarrow$ T $\rightarrow$ R+D} & 2.44 & \underline{3.88} & \textbf{2.12} & \textbf{8.28} & \textbf{4.06} & \textbf{13.58} \\ 
            \hline
        \end{tabular}
        \vskip -2.5em
    \end{table}

    \begin{table}[h!]
        \caption{ \textbf{Quantitative results upon GMFlow flow estimators with different training settings (cf. Section~\ref{sec:train_dataset}).}}
        \label{table:gmflow results}
        \vspace{-.5em}
        \centering
        \begin{tabular}{ccccccccccc}
            \hline
            \multirow{2}{*}{Datasets} & \multicolumn{2}{c}{Sintel} & \multicolumn{2}{c}{KITTI-12} & \multicolumn{2}{c}{KITTI-15} \\
            \cline{2-3} \cline{4-5} \cline{6-7}
              & clean & final & EPE & F1-all & EPE & F1-all \\
            \hline
            \texttt{C} & 3.23 & \underline{4.43} & 8.73 & 47.10 & 17.82 & 56.15 \\
            \texttt{C $\rightarrow$ T} & \textbf{1.50} & \textbf{2.96} & 5.09 & 25.75 & 11.60 & 35.52  \\
            \texttt{R+D} & 3.58   & 4.88   & \textbf{4.21} & \underline{21.63} & \underline{9.80} & \underline{33.95} \\
            \texttt{C $\rightarrow$ T $\rightarrow$ R+D} & \underline{3.01} & 4.60 & \underline{4.33} & \textbf{19.60} & \textbf{8.66} & \textbf{28.78} \\
            \hline
        \end{tabular}
        \vskip -1.5em
    \end{table}
    In Figure \ref{fig:kitti results} we provide some qualitative examples, where we see that the models trained on only synthetic datasets though provide correct object shape but may exhibit glaring errors in flow direction, while the models trained on diverse and challenging real-world datasets (attributed to our general flow generation and lateral geometric augmentation) show less artifacts and errors, thus achieving better performance.

\subsection{Ablation Study}
    \noindent \textbf{Model Designs.} We conduct an investigation upon the model designs in our proposed framework: virtual disparity, lateral geometric augmentation, and auxiliary classifier. Please note that here we specifically only adopt DIML dataset to perform model training in order to exclude the benefit of our unifying various depth datasets and better focus on the contributions of our model designs. Moreover, as virtual ego-motion has been introduced in \cite{aleotti2021learning}, we thus do not consider it in our ablation study. Furthermore, though DIML dataset itself is already a stereo dataset, we can still leverage different scaling factors for baseline $B$ and focal length $f$ to introduce difference virtual disparities. From Table \ref{table:DIML ablations} which summarizes the ablation results on KITTI-15 dataset, we observe that: \textbf{1)} the model variant that excludes all our designs performs the worse, while introducing our virtual disparity to enrich the depth variance helps to boost the performance; \textbf{2)} the further introduction of lateral geometric augmentation provides the significant improvement thanks to its providing more diverse and challenging optical flow samples for learning; \textbf{3)} with adopting our proposed auxiliary classifier $C$, the full model exhibits further advance, thus verifying the contribution and efficacy of the corresponding novel objective $\mathcal{L}_C$.   
    

    \begin{table}[h!]
    \vspace{-.5em}
        \renewcommand{\arraystretch}{1.2}
        \caption{ \textbf{Ablation study for model designs}. Evaluation is based on KITTI-15 dataset with two flow estimator backbones (GMFlow and RAFT) being trained on DIML dataset.}
        \label{table:DIML ablations}
        \vspace{-.5em}
        \centering
        \scalebox{.9}{\begin{tabular}{ccccc|cccc}
            \hline
              virtual & lateral geometric & auxiliary & \multicolumn{2}{c}{GMFlow} & \multicolumn{2}{c}{RAFT} \\
            \cline{4-5}\cline{6-7}
              disparity & augmentation & classifier & EPE & F1-all & EPE & F1-all \\
            \hline
             \XSolid & \XSolid & \XSolid & 11.25 & 39.8 & 6.08 & 16.74  \\
             \Checkmark & \XSolid & \XSolid & 12.31 & 39.91 & 5.76 & 16.03  \\
             \Checkmark & \Checkmark & \XSolid & \underline{10.97} & \underline{35.44} & \underline{4.64} & \underline{15.83} \\
             \Checkmark & \Checkmark & \Checkmark & \textbf{10.94} & \textbf{34.05} & \textbf{4.52} & \textbf{15.36}  \\
            \hline
        \end{tabular}}
        \vskip -1em
    \end{table}
    Moreover, we conduct a further study upon the impact caused by the accuracy of auxiliary classifier $C$, in which we observe that a stronger $C$ (with its accuracy on identifying augmentation type to be 0.80) can contribute to better training of the flow estimator (i.e. 4.52 EPE and 15.36 F1-all in KITTI-15 for the RAFT model trained on \texttt{D}) in comparison to the weaker $C$ (with 0.69 accuracy to identify augmentation type, resulting in 5.10 EPE and 15.61 F1-all). Please kindly refer to our supplementary video for more ablation studies, implementation details, and qualitative results.
    \section{Conclusion}

We propose a novel framework to well unify various supervised depth estimation datasets, including both monocular and stereo ones, for synthesizing the real-world optical flow training set with groundtruth annotations. With further introducing the challenging optical flow training samples by our proposed lateral geometric augmentation and building a novel objective function based on our proposed auxiliary classifier, the learning of optical flow estimator is largely benefited to achieve the superior performance with respect to the state-of-the-art baseline across various experiments.

	
    \addtolength{\textheight}{-0cm}   
    
    
    \bibliographystyle{IEEEtran}
    \bibliography{literatur}
	
\end{document}